%% file: main.tex
\documentclass[submission,copyright,creativecommons]{eptcs}
\pdfoutput=1
\usepackage{breakurl}             
\usepackage{underscore}           

\usepackage{multicol}
\usepackage{tikz}
\usepackage{subfig}
\usetikzlibrary{petri,arrows,backgrounds,matrix,automata,positioning,shapes,shadows,patterns,fit,calc}
\tikzset{
        ->, >=stealth, node distance=1.5cm, every state/.style={thick, minimum size = 0pt}, 
	initial text=$ $,
}

\title{ASP-Based Declarative Process Mining\\
(Extended Abstract)}
\author{Francesco Chiariello
\institute{DIAG}
\institute{Sapienza University of Rome}
\email{chiariello@diag.uniroma1.it}
\and
Fabrizio Maria Maggi
\institute{KRDB}
\institute{Free University of Bozen-Bolzano}
\email{maggi@inf.unibz.it}
\and
Fabio Patrizi
\institute{DIAG}
\institute{Sapienza University of Rome}
\email{patrizi@diag.uniroma1.it}
}

\def\titlerunning{ASP-Based Declarative Process Mining}
\def\authorrunning{F. Chiariello, F.M. Maggi, F. Patrizi}

\input{macros}

\begin{document}
\maketitle

\begin{abstract}
We propose Answer Set Programming (ASP) as an approach for modeling and solving problems from the area of Declarative Process Mining (DPM). We consider here three classical problems, namely, Log Generation, Conformance Checking, and Query Checking. These problems are addressed from both a \emph{control-flow} and a \emph{data-aware} perspective. The approach is based on the representation of process specifications as (finite-state) automata. Since these are strictly more expressive than the de facto DPM standard specification language
\declare, more general specifications than those typical of DPM can be handled, such as formulas in linear-time  temporal logic over finite traces.
(Full version available in the Proceedings of the 36th AAAI Conference on Artificial Intelligence~\cite{AAAI22}).
\end{abstract}

\section{Contribution}

Process Mining (PM) \cite{DBLP:books/sp/Aalst16} is a research area concerned with analyzing \emph{event logs} stored by (enterprise) information systems, with the aim of better understanding the (business) processes that produce the logs and how they are enacted in the organization owning the systems. A \emph{process} can be thought of as a set of event sequences that reach a desired goal, whose observed sequences, called \emph{process traces}, constitute the event log. Different formalisms have been proposed by the Business Process Management (BPM) community for modeling processes, including Petri nets and  BPMN \cite{DBLP:books/daglib/Weske12}.
In Declarative Process Mining (DPM) \cite{DBLP:journals/ife/AalstPS09} process models are specified declaratively, in a constraint-based fashion, i.e., as sets of constraints, which must be satisfied during the process execution. This type of specifications is particularly suitable for knowledge intensive processes \cite{DBLP:journals/jodsn/CiccioM015} that include a large variety of behaviors and can be more compactly represented under an open world assumption (all behaviors that do not violate the constraints are allowed). Typical process modeling languages for specifying processes in a declarative way are \declare~\cite{DBLP:journals/ife/AalstPS09} and~\LTLf~\cite{DBLP:conf/ijcai/GiacomoV13}.

We consider three classical problems from (D)PM:
\begin{itemize}
    \item \emph{Log generation} \cite{DBLP:conf/bpm/SkydanienkoFGM18}, i.e., the problem of generating a set of traces
    of a given length, compliant with a process model;
    \item \emph{Conformance checking}  \cite{DBLP:journals/eswa/BurattinMS16}, i.e., the problem of checking whether the traces of a log are compliant with a process model;
    \item \emph{Query checking} \cite{DBLP:conf/otm/RaimCMMM14}, i.e., the problem of finding properties of (the process associated to) a log by checking its conformance with the instantiation of candidate template formulas (also called \emph{queries}).
\end{itemize}
\noindent
In these problems, \emph{traces} are finite sequences of \emph{events}, which represent activity executions. \emph{Activities} model the atomic operations a process may perform and include \emph{attributes}, which events instantiate at execution time with specific values.

The \emph{control-flow} perspective focuses only on process activities, and the specification languages used in this setting include \declare or the more general \LTLf. In the \emph{data-aware} perspective, instead, also attributes and their values, i.e., the data, are taken into account, and richer logics are used, such as \mpdeclare \cite{DBLP:journals/eswa/BurattinMS16} or \LTLf with local conditions (\LLTLf); this is the setting adopted in the paper.

To solve the problems of our interest, we exploit the fact that, analogously to the case of \LTLf, for every \LLTLf formula, there exists a finite-state automaton (FSA) accepting exactly the traces that satisfy the formula. While every problem has its own specific features, they all share some common parts. 
As a consequence, the encoding approach of each problem includes the following steps:
\begin{itemize}
    \item construction of the FSAs corresponding to  each \LLTLf input specification;
    \item definition of an encoding schema for FSAs, as a set of ASP rules;
    \item definition of an encoding schema for traces, as a set of ASP facts;
    \item definition of ASP rules to simulate the execution of an FSA on a trace;
    \item definition of ASP generation and test rules (when needed) for the specific problem.
\end{itemize}
For Log Generation, the generation rules require that each position (up to the specified length) contains exactly one activity whose attributes are assigned one value from their respective domain. Then, test rules check whether the FSAs corresponding to the process model accept the generated trace.
Conformance Checking is even simpler since traces are given. To deal with many traces, we associate an index to each of them and then add a new argument to predicates, which represents the index. 
Then, for every pair of index and FSA, we check  
whether the FSA accepts the trace identified by the index. 
For Query Checking, an instantiation of the variables in the template formula is first guessed, in order to obtain a proper \LLTLf formula. Then, all traces are checked against the FSA corresponding to the so-obtained formula. Special care is required in the ASP encoding to deal with formula variables; to this end, we introduce suitable predicates modeling their instantiation.

For all problems, we provide the ASP encoding, an experimental evaluation and, when possible, a comparison with SoA tools. The obtained results show that our approach considerably outperforms the SoA \emph{\mpdeclare Log Generator} \cite{DBLP:conf/bpm/SkydanienkoFGM18} based on Alloy \cite{jackson2012software} (in turn based on SAT solvers), which does not exploit the FSA representation of formulas.
Slightly worse results are obtained for Conformance Checking wrt the \emph{\declare Analyzer} \cite{DBLP:journals/eswa/BurattinMS16}, which is however tailored to \declare and, differently from the ASP-based approach, cannot be used to check general \LTLf~formulas. 
As to Query Checking, this is the first solution in a data-aware setting, thus, while the experiments show the feasibility of the approach, no comparison with other tools is carried out.
The ASP implementation of the approach is not optimized, as the focus of the work is on feasibility rather than performance, so further improvements are possible.


As an example of FSA encoding, consider the (control-flow) \emph{response}
 constraint $\varphi=\always(a\limp \eventually b)$ saying that ``whenever $a$ occurs, it must be eventually followed by $b$''.\footnote{Differently from the generic traces usually considered with {\LTLf}, here two activities cannot be true at the same time.} 
 The FSA shown in Figure \ref{fig:example} accepts all and only the traces that satisfy $\varphi$. The corresponding ASP representation is reported next to the automaton. Predicates {\init} and {\acc} are used to model initial and accepting states, respectively. Predicate {\trans} models state transitions and are labeled each with a unique integer, identifying the transition. Finally, predicate {\hold} models which transitions are enabled by the trace event at time $T$. The automaton execution on the input trace is then simulated by reading the trace event by event, with the automaton starting in its initial state, and then following, at every step, the transition enabled by the read event. If the formulae labeling the transitions express more complex properties involving data conditions, these can be modeled in the body of the rules for predicate {\hold}.
For example, the following rule expresses that transition 1 is enabled at time $T$ if $a$ occurs at $T$, with a value less than $5$ for attribute $number$:
$\hold(1,T) \larr \trace(a,T), \hvalue(number,V,T), V<5.$
\begin{figure}
\centering
\subfloat{{\begin{tikzpicture}
\node[state, initial, accepting] (s0) {$s_0$};
\node[state, right of=s0, xshift=.5cm] (s1) {$s_1$};
\draw
(s0) edge[loop above] node{$\lnot a$} (s0)
    	(s0) edge[bend left, above] node{$a$} (s1)
    	(s1) edge[loop above] node{$\lnot b$} (s1)
    	(s1) edge[bend left, above] node{$b$} (s0);
\end{tikzpicture}}}%
\hspace{1cm}
\subfloat{
\begin{small}
$\begin{array}{l}
\init(s_0).\\
\acc(s_0).\\
\trans(s_0,1,s_1).\\
\hold(1,T) \larr \trace(a,T).\\
\trans(s_1,2,s_0).\\
\hold(2,T) \larr \trace(b,T).\\
\trans(s_0,3,s_0).\\
\hold(3,T) \larr \naf \hold(1,T), \timep(T).\\
\trans(s_1,4,s_1).\\
\hold(4,T) \larr  \naf \hold(2,T), \timep(T).\\
\end{array}$
\end{small}
}
\caption{FSA of formula $\varphi=\always(a\limp \eventually b)$ and its ASP representation.\label{fig:example}}
\end{figure}

By showing how to handle \LTLf specifications, namely by exploiting the FSA representation for reducing problems to reachability of accepting states, we have paved the way for the use of ASP as a solving approach to DPM problems and, potentially, to all problems involving temporal specifications over finite traces.

\paragraph{Acknowledgements}
Work partly supported by: ERC Advanced Grant WhiteMech (No. 834228),
EU ICT-48 2020 project TAILOR (No. 952215), 
Sapienza Project DRAPE, 
UNIBZ project CAT.

\bibliographystyle{eptcs}
\bibliography{biblio}

\end{document}

%% file: macros.tex
\usepackage{xspace}

\newcommand{\LTLf}{{\sc ltl}$_f$\xspace}
\newcommand{\LLTLf}{{\sc l-ltl}$_f$\xspace}

\newcommand{\declare}{{\sc declare}\xspace}
\newcommand{\mpdeclare}{{\sc mp-declare}\xspace}

\newcommand{\always}{\mathbf{G}}
\newcommand{\eventually}{\mathbf{F}}

\newcommand{\naf}[1]{\nafop~{#1}}
\newcommand{\nafop}{\mathtt{not}}

\newcommand{\limp}{{\rightarrow}}
\newcommand{\larr}{\leftarrow}

\newcommand{\hvalue}{\texttt{has\_val}}
\newcommand{\timep}{\texttt{time}}
\newcommand{\trace}{\texttt{trace}}
\newcommand{\init}{\texttt{init}}
\newcommand{\acc}{\texttt{accepting}}
\newcommand{\trans}{\texttt{trans}}
\newcommand{\hold}{\texttt{hold}}

%% file: main.bbl
\begin{thebibliography}{10}
\providecommand{\bibitemdeclare}[2]{}
\providecommand{\surnamestart}{}
\providecommand{\surnameend}{}
\providecommand{\urlprefix}{Available at }
\providecommand{\url}[1]{\texttt{#1}}
\providecommand{\href}[2]{\texttt{#2}}
\providecommand{\urlalt}[2]{\href{#1}{#2}}
\providecommand{\doi}[1]{doi:\urlalt{http://dx.doi.org/#1}{#1}}
\providecommand{\eprint}[1]{arXiv:\urlalt{https://arxiv.org/abs/#1}{#1}}
\providecommand{\bibinfo}[2]{#2}

\bibitemdeclare{book}{DBLP:books/sp/Aalst16}
\bibitem{DBLP:books/sp/Aalst16}
\bibinfo{author}{Wil M.~P. \surnamestart van~der Aalst\surnameend}
  (\bibinfo{year}{2016}): \emph{\bibinfo{title}{Process Mining - Data Science
  in Action, Second Edition}}.
\newblock \bibinfo{publisher}{Springer}, \doi{10.1007/978-3-662-49851-4}.

\bibitemdeclare{article}{DBLP:journals/ife/AalstPS09}
\bibitem{DBLP:journals/ife/AalstPS09}
\bibinfo{author}{Wil M.~P. \surnamestart van~der Aalst\surnameend},
  \bibinfo{author}{Maja \surnamestart Pesic\surnameend} \&
  \bibinfo{author}{Helen \surnamestart Schonenberg\surnameend}
  (\bibinfo{year}{2009}): \emph{\bibinfo{title}{Declarative workflows:
  Balancing between flexibility and support}}.
\newblock {\sl \bibinfo{journal}{Comput. Sci. Res. Dev.}}
  \bibinfo{volume}{23}(\bibinfo{number}{2}), pp. \bibinfo{pages}{99--113},
  \doi{10.1007/s00450-009-0057-9}.

\bibitemdeclare{article}{DBLP:journals/eswa/BurattinMS16}
\bibitem{DBLP:journals/eswa/BurattinMS16}
\bibinfo{author}{Andrea \surnamestart Burattin\surnameend},
  \bibinfo{author}{Fabrizio~Maria \surnamestart Maggi\surnameend} \&
  \bibinfo{author}{Alessandro \surnamestart Sperduti\surnameend}
  (\bibinfo{year}{2016}): \emph{\bibinfo{title}{Conformance checking based on
  multi-perspective declarative process models}}.
\newblock {\sl \bibinfo{journal}{Expert Syst. Appl.}} \bibinfo{volume}{65},
  \doi{10.1016/j.eswa.2016.08.040}.

\bibitemdeclare{inproceedings}{AAAI22}
\bibitem{AAAI22}
\bibinfo{author}{Francesco \surnamestart Chiariello\surnameend},
  \bibinfo{author}{{Fabrizio Maria} \surnamestart Maggi\surnameend} \&
  \bibinfo{author}{Fabio \surnamestart Patrizi\surnameend}
  (\bibinfo{year}{2022}): \emph{\bibinfo{title}{{ASP}-Based Declarative Process
  Mining}}.
\newblock In: {\sl \bibinfo{booktitle}{Proc.~of the 36th AAAI Conference on
  Artificial Intelligence, {AAAI} 2022}}.

\bibitemdeclare{inproceedings}{DBLP:journals/jodsn/CiccioM015}
\bibitem{DBLP:journals/jodsn/CiccioM015}
\bibinfo{author}{Claudio~Di \surnamestart Ciccio\surnameend},
  \bibinfo{author}{Andrea \surnamestart Marrella\surnameend} \&
  \bibinfo{author}{Alessandro \surnamestart Russo\surnameend}
  (\bibinfo{year}{2015}): \emph{\bibinfo{title}{Knowledge-Intensive Processes:
  Characteristics, Requirements and Analysis of Contemporary Approaches}}.

\bibitemdeclare{inproceedings}{DBLP:conf/ijcai/GiacomoV13}
\bibitem{DBLP:conf/ijcai/GiacomoV13}
\bibinfo{author}{Giuseppe \surnamestart {De Giacomo}\surnameend} \&
  \bibinfo{author}{Moshe~Y. \surnamestart Vardi\surnameend}
  (\bibinfo{year}{2013}): \emph{\bibinfo{title}{Linear Temporal Logic and
  Linear Dynamic Logic on Finite Traces}}.
\newblock In: {\sl \bibinfo{booktitle}{Proc. of the 23rd Int. Joint Conf. on
  Artificial Intelligence}}, \bibinfo{publisher}{{IJCAI/AAAI}}.

\bibitemdeclare{book}{jackson2012software}
\bibitem{jackson2012software}
\bibinfo{author}{Daniel \surnamestart Jackson\surnameend}
  (\bibinfo{year}{2012}): \emph{\bibinfo{title}{Software Abstractions: logic,
  language, and analysis}}.
\newblock \bibinfo{publisher}{MIT Press}.

\bibitemdeclare{inproceedings}{DBLP:conf/otm/RaimCMMM14}
\bibitem{DBLP:conf/otm/RaimCMMM14}
\bibinfo{author}{Margus \surnamestart R{\"{a}}im\surnameend},
  \bibinfo{author}{Claudio~Di \surnamestart Ciccio\surnameend},
  \bibinfo{author}{Fabrizio~Maria \surnamestart Maggi\surnameend},
  \bibinfo{author}{Massimo \surnamestart Mecella\surnameend} \&
  \bibinfo{author}{Jan \surnamestart Mendling\surnameend}
  (\bibinfo{year}{2014}): \emph{\bibinfo{title}{Log-Based Understanding of
  Business Processes through Temporal Logic Query Checking}}.
\newblock In: {\sl \bibinfo{booktitle}{On the Move to Meaningful Internet
  Systems: {OTM} 2014 Conferences}}, \doi{10.1007/978-3-662-45563-0_5}.

\bibitemdeclare{inproceedings}{DBLP:conf/bpm/SkydanienkoFGM18}
\bibitem{DBLP:conf/bpm/SkydanienkoFGM18}
\bibinfo{author}{Vasyl \surnamestart Skydanienko\surnameend},
  \bibinfo{author}{Chiara~Di \surnamestart Francescomarino\surnameend},
  \bibinfo{author}{Chiara \surnamestart Ghidini\surnameend} \&
  \bibinfo{author}{Fabrizio~Maria \surnamestart Maggi\surnameend}
  (\bibinfo{year}{2018}): \emph{\bibinfo{title}{A Tool for Generating Event
  Logs from Multi-Perspective Declare Models}}.
\newblock In: {\sl \bibinfo{booktitle}{Proceedings of the Dissertation Award,
  Demonstration, and Industrial Track at {BPM} 2018}}.

\bibitemdeclare{book}{DBLP:books/daglib/Weske12}
\bibitem{DBLP:books/daglib/Weske12}
\bibinfo{author}{Mathias \surnamestart Weske\surnameend}
  (\bibinfo{year}{2012}): \emph{\bibinfo{title}{Business Process Management -
  Concepts, Languages, Architectures, 2nd Edition}}.
\newblock \bibinfo{publisher}{Springer}, \doi{10.1007/978-3-642-28616-2}.

\end{thebibliography}
